# Small Aerial Target Detection for Airborne Infrared Detection Systems using LightGBM and Trajectory Constraints

Xiaoliang Sun*, Liangchao Guo*, Wenlong Zhang, Zi Wang and Qifeng Yu

*Abstract*—Factors, such as rapid relative motion, clutter background, etc., make robust small aerial target detection for airborne infrared detection systems a challenge. Existing methods are facing difficulties when dealing with such cases. We consider that a continuous and smooth trajectory is critical in boosting small infrared aerial target detection performance. A simple and effective small aerial target detection method for airborne infrared detection system using light gradient boosting model (LightGBM) and trajectory constraints is proposed in this article. First, we simply formulate target candidate detection as a binary classification problem. Target candidates in every individual frame are detected via interesting pixel detection and a trained LightGBM model. Then, the local smoothness and global continuous characteristic of the target trajectory are modeled as short-strict and long-loose constraints. The trajectory constraints are used efficiently for detecting the true small infrared aerial targets from numerous target candidates. Experiments on public datasets demonstrate that the proposed method performs better than other existing methods. Furthermore, a public dataset for small aerial target detection in airborne infrared detection systems is constructed. To the best of our knowledge, this dataset has the largest data scale and richest scene types within this field.

*Index Terms*—Airborne, infrared detection system, light gradient boosting model (LightGBM), small aerial target, trajectory constraint.

## I. Introduction

Compared to optical or radar detection, infrared detection has the advantages of all-day and all-weather operation, high resolution, and strong concealment simultaneously. Thus, the infrared detection systems have been widely used in air platform early warning and guidance [1]. Small infrared target detection is the key technology in such detection systems. However, for the airborne infrared detection systems, which are subject to rapid relative motion and cluttered backgrounds, small aerial target detection is still a challenge.

Manuscript received 16 May 2021; revised 11 July 2021 and 6 September 2021; accepted 23 September 2021. Date of publication 27 September 2021; date of current version 13 October 2021. This work was supported in part by the Hunan Provincial Natural Science Foundation of China under Grant 2019JJ50732 and in part by the National Natural Science Foundation of China under Grant 62003357. (Xiaoliang Sun and Liangchao Guo contributed equally to this work.) (Corresponding author: Xiaoliang Sun.)

The authors are with the College of Aerospace Science and Engineering, National University of Defense Technology, Changsha 410073, China (e-mail: alexander_sxl@nudt.edu.cn; guoliangchao19@nudt.edu.cn; wenlong@nudt.edu.cn; wangzitju@163.com; qifeng_yu1958@126.com).

Digital Object Identifier 10.1109/JSTARS.2021.3115637

Small infrared target detection has been studied for a long time. The information contained in a single frame or multiple successive frames is used in small infrared target detection. Accordingly, existing methods can be categorized into single frame based methods and multiple successive frames based methods [1]. Single frame based methods detect the small infrared target mainly using the differences between the target and background. The small infrared target is often modeled as a spot target of isotropic distribution [3, 4]. The distribution is determined based on the point spread function of the imaging system or simply approximated by Gaussian distribution. The background characteristic [5, 7] and the local contrast [15, 16] are also widely used in single frame based infrared small target detection. Such methods are efficient and easy to implement. However, cues provided by a single frame may be inadequate for robust small infrared target detection, especially in cases wherein the targets are extremely weak and the backgrounds are cluttered. Temporal cues contained in multiple successive frames, e.g., the high correlation of background and the continuity of the target, are important for robust small infrared target detection [1]. Multiple successive frames based methods boost the performance of small target detection by associating multiple images [35, 42]. However, the adoption of temporal cues increases the computational complexity in existing methods. Additionally, such methods cannot handle rapidly changing backgrounds well. In conclusion, existing methods, including single frame based and multiple successive frames based methods, have trouble in detecting small aerial targets for airborne infrared detection systems.

Multiple successive frames-based methods perform better than single frame-based methods generally. The fact indicates that those temporal cues are important in robust small infrared target detection, especially for complex cases. We analyze a large number of small aerial target image sequences captured by an airborne imaging platform. The backgrounds are varied, e.g., clouds, vegetation, water, and buildings. Cluttered backgrounds may not show high correlation within image sequences, especially for cases in which the background is changing rapidly due to the motion of the airborne infrared detection system. In this situation, the cues of the backgrounds are inappropriate choices for robust small infrared aerial target detection (SIATD). However, we find that true targets exhibit continuous and smooth long trajectories while the clutter does not. Based on this fact, this article tackles the challenge of small aerial target detection for airborne infrared detection systems by using the light gradient boosting model (LightGBM) [2] and trajectory constraints innovatively. Spatial and temporal cues within image sequences are used in a simple and effective way. Spatial cues are used in target candidate detection from each single frame. Target candidate detection is simply formulated as a binary classification problem. LightGBM is firstly introduced in infrared small target detection in this article. The proposed method detects the target candidates from each frame using interesting pixel detection and a trained LightGBM model. Temporal cues are used in detecting the true target from numerous target candidates. A simple piecewise uniform motion model is used to approximate the continuous and smooth target trajectory. We use a short-strict constraint to preserve the local linearity of the target trajectory strictly, and a long-loose constraint to extend the continuous target trajectory to the fullest extent. Based on the short-strict and long-loose constraints, the true targets are detected via trajectory segment growth and merging effectively. We evaluate the proposed method on public datasets (SIRST [39] and SIATD) and compare it with existing methods. Experimental results indicate that our method can detect small infrared aerial targets under a cluttered background robustly and achieve better performance than existing methods.

A high-quality dataset is of considerable significance for small infrared aerial target detection research. However, to the best of our knowledge, there is no publicly available dataset for small aerial target detection for airborne infrared detection systems at present. In this study, a large-scale dataset for small infrared aerial target detection is built and made available publicly (https://small-infrared-aerial-target-detection.grand-challenge.org/). It contains the largest data scale and the richest scene types in this research field thus far.

The main contributions of this paper are as follows:
1) Target trajectory constraints are modeled as short-strict and long-loose constraints innovatively. A simple and effective small infrared aerial target detection method using LightGBM and trajectory constraints is proposed. It demonstrates superior performance compared to existing methods.
2) A publicly available high-quality dataset for small aerial target detection for airborne infrared detection systems is built.

The rest of the article is organized as follows: Section II summarizes the related works. In Section III, we analyze the small aerial target's trajectory characteristics in image sequences captured using airborne infrared imaging equipment. Details of the proposed method are presented in Section IV. Section V gives an elaborated description of the dataset. Experimental results are given in Section VI. Finally, Section VII concludes the article.

## II. Related works

As mentioned above, existing small infrared target detection methods can be classified into single frame-based and multiple frames-based methods. We also summarize the most recently published representative convolutional neural network (CNN) related methods separately.

### A. Single frame based methods

Single-frame-based methods detect small infrared targets from a single image. They mainly use the difference between the target and the background. Based on the characteristics used, such methods can be further subdivided as follows.

*1) Target Appearance Characteristic-Based Methods:* These methods detect small infrared targets mainly based on the target's appearance. In remote imaging, a small infrared target is often modeled as a spot in the image. Its appearance distribution is determined by the imaging system. Moradi *et al.* [3] modeled the spot target using the point spread function of the imaging system. Zhang *et al.* [4] adopted isotropic distributions to approximate the target. Although such methods are simple and efficient, they perform poorly in cluttered backgrounds.

*2) Background Large Spatial Spread Characteristic-Based Methods:* Based on the remote imaging and infrared imaging characteristics, researchers model the background approximately using common or uniform components. Compared to the background, the small target has a small spatial spread. The small target is detected by subtracting the estimated background. Random walker and facet kernel filter are combined in [5] to detect a small infrared target. Similar to random walker, Huang et al. [6] proposed a structure adaptive clutter suppression method named chain-growth filtering. Gao *et al.* [7] modeled the background using Infrared Patch-Image (IPI) reconstruction. In order to eliminate the negative influences of the strong edges, Dai et al. adopted non-negative constraints in [8]. A local and global priors Reweighted Infrared Patch Tensor (RIPT) model was described in [9]. Zhang *et al.* [10] modified IPI by introducing a three dimension tensor model. Xue *et al.* [11] introduced multiple sparse constraints in reconstruction. Lv *et al.* [12] proposed an efficient online update method to improve the traditional dictionary learning algorithm for small infrared target detection. In order to eliminate the negative influence of complex backgrounds, Zhou *et al.* [13] introduced $l_{1/2}$–metric and dual-graph regularization in sparse component modeling. Guan *et al.* [14] improved the IPT model using a non-convex tensor rank surrogate merging tensor nuclear norm and the Laplace function. This type of method performs well when the background satisfies the large spatial spread assumption but they cannot handle cluttered backgrounds satisfactorily. Furthermore, the background modeling is often time consuming.

*3) Local Characteristic Difference-Based Methods:* The small infrared target and background belong to different components in images and exhibit different characteristics. The characteristic difference in the target centered local region is used to detect a small infrared target. The two widely-used structures of the local region are shown in Fig. 1 [15].

The gray statistical information, i.e., mean and variance, are computed for each part. The differences between them are used as cues for small target detection. Modifications toward interferences, e.g., isolated spot noise, strong background edge, clutter, and size changes, are made to improve performance. Related works include Improved Local Contrast Measure (ILCM) [16], Multiscale Local Contrast Measure (MLCM) [17], Weighted Local Difference Measure (WLDM) [18],

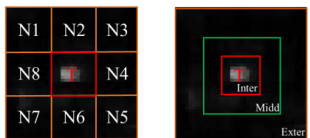

Fig. 1. Structure of target centered local region. (a) Eight neighbors, (b) Concentric rectangles (From inside to outside: the internal (Inter), middle (Midd), and external (Exter) area).

Multiscale Gray and Variance Difference (MGVD) [15]. Zhao et al. [19] extracted contrast information of small infrared target in the max-tree and min-tree and proposed a novel detection method based on multiple morphological profiles. Li et al. [20] use a local steering kernel to estimate the local intrinsic structure of the local image patch. These methods suffer from a high false-alarm rate for cluttered backgrounds.

### B. Multiple Frames-Based Methods

Single frame based methods are easy to implement. However, they have trouble in detecting small infrared targets under a cluttered background. Multiple frames based methods introduce temporal cues in target detecting via association between successive frames. We subdivide related works into pixel and patch association-based methods.

*1) Single Pixel Association-Based Methods:* Such methods obtain the temporal gray distribution for each individual pixel by associating multiple successive frames. The target is detected by combining the priors of the target or depending on the differences between the temporal gray distributions directly. Wu et al. [21] detected a small infrared target by analyzing the correlation between temporal profiles using a kernel algorithm. In [22], researchers detected the small target robustly by combining the movement and appearance cues. Marco et al. [23] proposed a generalized likelihood ratio test-based method for small target detection in an oceanic background. Nichols et al. [24] modified to the method by proposing an iterative solution for the max likelihood estimation. Sun et al. [1] proposed an efficient energy accumulation method for small target detection based on dynamic programming analysis. James et al. [25] designed the small target detection metric using the Markov model theory. Fan et al. [26] adopted passion distribution in energy accumulation for small infrared target detection. Kwan et al. [27] adopted optical flow techniques to enhance small moving infrared target detection performance, especially for low-quality and long-range infrared videos. Single pixel association-based methods are sensitive to cluttered backgrounds and isolated spot noise.

*2) Image Patch Association-Based Methods:* Compared to a single pixel, an image patch contains more meaningful cues that can improve the robustness of small target detection. Li *et al*. [28] constructed image patches sparsely and combined a particle filter to detect small targets from an image sequence. In order to eliminate the negative influence of clutter in image patches, Qian *et al*. [29] adopted a guided filter and Gaussian weight to suppress the cluttered background. Dong *et al*. [30] first estimated the motion of the imaging platform using image patch correlation and then detected the true target trajectory based on the trajectory continuity. Li *et al*. [31] enhanced the small infrared target via saliency analysis based on motion and appearance. Lv *et al*. [32] suppressed the highly correlated backgrounds within successive frames and then detected the target based on local gray distribution. Ren *et al*. [33] treated the small infrared target as noise and detected it using a denoising algorithm. Gao *et al*. [34] expended IPI to multiple frames. The background is modeled as a low rank matrix and a mixture Gaussian model is adopted to model the target. The target is then detected from the reconstructed result. Liu *et al*. [35] proposed a non-convex tensor low-rank approximation method to estimate the clutter background accurately. Sun et al. [36] modified the method by using a non-i.i.d mixture of Gaussian with modified flux density. The spatio-temporal tensor model is adopted to model the background in [37, 38]. The authors of [39] proposed a novel robust principal component analysis based on weighted Schatten-p norm to improve the accuracy of background estimation. Multiple subspace learning is adopted to modify [39] in [40]. Taking edge and corner into consideration, Zhang *et al*. [41] proposed a novel spatial-temporal tensor model to detect infrared small target. Motivated by human visual perception, Li *et al*. [42] proposed a novel spatio-temporal saliency approach for dim moving target detection. Image patch association methods perform well for cases in which the backgrounds are stable but cannot handle rapidly changing backgrounds satisfactorily. In addition, the methods are often complex and cannot meet the needs of real-time applications.

### C. CNN Related Methods

Convolutional Neural Networks (CNNs) have achieved considerable success in many vision applications, e.g., detection, classification, and segmentation. They have also been used in small infrared target detection. Many deep networks mainly rely on object-appearance-centered feature representation. However, given the scarcity of target intrinsic characteristics and the presence of clutter in the backgrounds, conventional deep networks easily fail in small infrared target detection [43, 44]. Furthermore, most convolutional networks attenuate the feature map size to learn high-level semantic features, which may result in the small infrared target being over-whelmed by cluttered backgrounds. To address this, Dai et al. [43] preserved and highlighted the small target feature by exploiting a bottom-up attentional modulation integrating the low-level features into the high-level features of deeper layers. The authors of [44] constructed Generative Adversarial Networks upon a U-Net to learn the features of small infrared targets and directly predict the intensity of targets.

Most existing CNN related works are single-frame-based methods. Temporal cues have not been adopted in the abovementioned works. They have difficulties in detecting small infrared aerial targets under cluttered backgrounds, especially for small and dim targets.

## III. ANALYZING SMALL INFRARED AERIAL TARGET'S TRAJECTORY CHARACTERISTICS

This article focuses on small aerial target detection for airborne infrared detection systems. In remote imaging, the target is often presented as a single spot in the image. Due to the motion of the imaging platform, the cluttered backgrounds in the image sequences change rapidly. Existing methods face difficulties in dealing with such cases. We have analyzed numerous small infrared aerial target image sequences. The movement of the aerial target follows the basic laws of physics. In general, the targets in successive multi-frames form continuous and smooth long trajectories. Fig. 2 presents two sample image sequences.

Sample sequence #1 and #2 each contain 200 frames. The changes in the background are caused by the imaging system motion. The target candidates in each frame are detected using the method presented in Section IV-A. We adopt the homograph transform to model the inter-frame movement. Inter-frame registration is performed based on feature point extraction and matching (see Section IV-B). We remap the target candidates within the sequence to the same coordinate system. The remapped results in Fig. 2 show that the true targets' trajectories are continuous and smooth where the noises do not form continuous trajectories. The characteristics of the target's trajectory have been used in small infrared target detection. Existing methods often make assumptions about the target motion, e.g., the target moves at a constant speed. Such assumptions may not hold in real-world applications, especially over a long time range. The targets in Fig. 2 move with changing velocity and form curved trajectories. Thus, the related methods cannot track the target well over a long time range. However, a long trajectory is important for robust small infrared aerial target detection. Therefore, related methods have limitations in detecting moving small infrared aerial target robustly in practice.

Although the strong assumptions mentioned above may not hold for long trajectories, the characteristics of continuity and smoothness hold for most targets' trajectories. Table I provides the trajectory candidates of various lengths formed by target candidates within successive frames of sample sequences #1 and #2. Details of the trajectory candidate generation are presented in Section IV-B.

Sequence #1 contains one small aerial target and #2 contains three. The targets may exit and re-enter the field of view. Thus, the true target trajectory number may be larger than the target number within the sequence. As presented in Table I, with the increase of the trajectory length, the number of false trajectories decreases. False target trajectories are almost eliminated at a length of 13. The results indicate that the extending of the trajectory length has considerable significance for robust SIATD.

Motivated by the analysis above, this article adopts LightGBM and target trajectory constraints to robustly detect small infrared aerial targets for airborne detection systems. Different from existing methods, this article adopts trajectory constraints in a simple but effective way. The proposed method finds the true target trajectory by linking target candidates directly, not by associating pixels or image patches as existing methods do. The trajectory constraints include the short-strict and long-loose constraints. Details of the proposed method are provided in the following section.

## IV. PROPOSED APPROACH

This article uses LightGBM and trajectory constraints in robust SIATD for airborne detection systems. The proposed method first extracts multiple spatial features and detects target candidates from each frame using interesting pixel detection and a trained LightGBM model. Then, in adopting trajectory constraints, we use the piecewise uniform motion model to approximate a continuous and smooth long target trajectory. Local linearity is guaranteed by a short-strict constraint on target motion. A long-loose constraint is proposed to link trajectory segments to form a continuous and smooth long trajectory. The true trajectories meeting the short-strict and long-loose constraints are finally detected.

### A. Target Candidate Detection for Each Frame

In remote imaging, the small aerial targets are often presented as spot targets in the image. However, the spot targets may be brighter or darker than their surroundings. In this study, the target candidate detection for each frame is formulated as a binary classification problem. For each pixel, we extract features in the local region centered on it. Then the trained LightGBM model takes the features as input and determines whether the pixel is a target candidate or not. In order to accelerate the target candidate detection, a positive and negative median filter is used to detect interesting pixels first.

*1) Interesting Pixel Detection:* The small infrared targets only correspond to a small portion of pixels in the images. Most pixels belong to the background. In order to detect the small infrared target efficiently, we first filter out the pixels that obviously correspond to the background. The remaining pixels are considered interesting pixels and subject to the following process. As mentioned above, the small infrared aerial target is often presented as a spot target. It may be brighter or darker than its neighbors. This article proposes a positive and negative

TABLE I
TRAJECTORY CANDIDATES OF DIFFERENT LENGTHS WITHIN SAMPLE SEQUENCES #1 AND #2.

| Sequence \ Length | 3 | 5 | 7 | 9 | 11 | 13 | 15 |
|---|---|---|---|---|---|---|---|
| #1 | 39 | 12 | 8 | 6 | 5 | 3 | 3 |
| #2 | 40 | 24 | 15 | 6 | 3 | 3 | 3 |

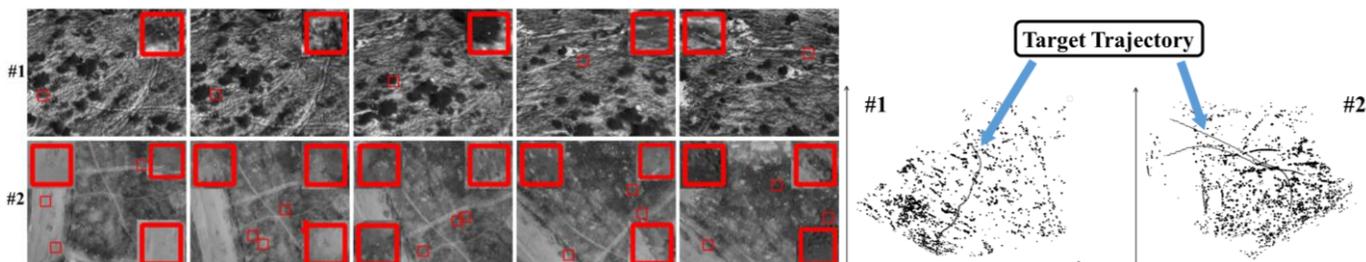

Fig. 2. Sample image sequences of small aerial infrared target. Right: Sample images from sequence #1 and #2. Targets are labeled by rectangles. Left: Remapped target candidates. The target candidates are remapped in the same coordinate system.

median filter to detect interesting pixels, as given in

$$\text{Label}(x,y) = \begin{cases} 1, & I(x,y) > (\text{median}((x,y))+k_1) \\ & \| I(x,y) < (\text{median}((x,y))+k_2) \\ 0, & \text{otherwise} \end{cases} \quad (1)$$

For the input image $I$, $\text{Label}(x,y)=1$ indicates that the pixel $(x,y)$ is an interesting pixel and $\text{Label}(x,y)=0$ denotes that it is not. $\text{median}((x,y))$ calculates the median value in a local region of a certain size centered on $(x,y)$. We set the parameters $k_1 > 0$ to select the brighter targets and $k_2 < 0$ for darker targets. It should be noted that each connected domain in Label is treated as a target candidate. The target candidate is denoted by the brightest/darkest pixel within the brighter/darker target. The values of $k_1$ and $k_2$ affects the number of target candidates selected from the input image. We deduce the values of $k_1$ and $k_2$ by varying from both sides to middle according to the required number of selected target candidates in practice.

*2) LightGBM Model:* Gradient boosting decision tree (GBDT) is an ensemble model of the decision trees, which are trained in sequence [45]. It learns the decision trees by fitting the residual error in each iteration. GBDT has been widely used in many machine learning tasks. However, its computational complexity is proportional to the number of instances and the number of features. This makes the traditional GBDT algorithm time-consuming when handling big data. Ke *et al*. [2] meet this challenge by proposing two techniques: Gradient-based One-side Sampling (GOSS) and Exclusive Feature Bundling (EFB). The modified GBDT algorithm is called LightGBM.

By combining GOSS and EFB, LightGBM accelerates the training process dramatically while achieving almost the same accuracy as the previous GBDT algorithms.

*3) Feature Extraction and Learning:* To train the LightGBM model, multiple spatial features are extracted for each pixel in the local region centered on it. For simplicity, we set the shape of the local region as a rectangle. In order to better capture the characteristics of the small infrared target in the image, seven features are computed from the local region, including kurtosis $\gamma_2$, skew $S_k$, entropy $H$, mean $\mu$, variance $\sigma^2$, maximum $v_{max}$, and minimum $v_{min}$. Let $L_{R_1 \times R_2}(x,y)$ denote the local rectangular region centered on the pixel $(x,y)$ of size $R_1 \times R_2$. We flatten the region into a vector $V = \{v_0, v_1, ......, v_{N-1}\}_{N=R_1 \times R_2}$. The definitions of the 7 spatial features are given in Equation (2). The mean and standard deviation are denoted as $\mu = E(v_i)$ and $\sigma$. $\mu_3 = E((v_i - \mu)^3)$ and $\mu_4 = E((v_i - \mu)^4)$ are the third and fourth central moment respectively. $p(\cdot)$ denotes the probability of the intensity value and it can be inferred from the intensity histogram of the input image.

$$\begin{aligned} \text{kurtosis:} \quad & \gamma_2 = \frac{\mu_4}{\sigma^4} - 3 \\ \text{skew:} \quad & S_k = \frac{\mu_3}{\sigma^3} \\ \text{entropy:} \quad & H(V) = -\sum_{v_i \in V} p(v_i) \log p(v_i) \\ \text{mean:} \quad & \mu = \frac{1}{N} \sum_{v_i \in V} v_i \\ \text{variance:} \quad & \sigma^2 = \frac{1}{N} \sum_{v_i \in V} (v_i - \mu) \\ \text{maximum:} \quad & v_{max} = \max_{v_i \in V}(v_i) \\ \text{minimum:} \quad & v_{min} = \min_{v_i \in V}(v_i) \end{aligned} \quad (2)$$

In the training dataset, the small infrared aerial targets are annotated. We take the pixels within the regions centered on the label positions of size $3 \times 3$ as positive samples. The remaining pixels in the images can be taken as negative samples. A seven-dimensional spatial feature vector, is calculated for each sample to train the LightGBM model. In order to detect targets of varied scales, we introduce a multiscale processing strategy.

*B. Target Detection Using Trajectory Constraints*

The target candidates in each frame are detected as mentioned above. Considering the remote imaging conditions in small aerial target detection for an airborne infrared detection system, the elevation variance of the background can be ignored compared to the imaging distance. Therefore, the homograph transform is adopted to model the inter-frame movement in this paper. The registration between successive frames is built via SURF [46] feature point extraction and matching. The transform parameters are solved via a RANSAC based robust method. Then we remap the target candidates in each frame to the coordinate of the first frame within the time window.

We intend to detect the true targets whose trajectories obey the short-strict and long-loose constraints. The trajectory of the small infrared aerial target captured by an airborne infrared detection system in Section III is analyzed. The analysis indicates that the true target forms a continuous and smooth long trajectory in the captured image sequence. The long trajectory can be used to distinguish targets from clutter robustly. The target's movement is modeled as a piecewise uniform motion. We impose the short-strict constraint, i.e., uniform motion, on the trajectory in a short time interval to eliminate the interference of clutter as much as possible. By contrast, we impose the long-loose constraint on the trajectory in a long time range to extend the length of the trajectory as much as possible. Other than the short-strict and long-loose constraints, we do not make strong assumptions about the speed of the small infrared aerial target as previous algorithms do. The trajectory synthetizing and validation include trajectory segment growth and merging. They are detailed as follows.

*1) Trajectory Segment Growth With Short-Strict Constraint:*
Trajectory segment growth links the target candidates in the current frame to the existing trajectory segments suitably. In this study, the target movement is modeled as a piecewise uniform motion, i.e., the short-strict constraint, in a short time interval. We set the short time interval as three successive frames. Trajectory segments grow under the short-strict constraint. We denote the existing trajectory segment set as $\{T^i\}_M$ and the target candidate set as $\{c_j^t\}_N$ in the current frame $t$.

As shown in Fig. 3, we take a sample trajectory segment $T^i = \{\cdots, n_{t-3}^i, n_{t-2}^i, n_{t-1}^i\}$ to detail the implementation of trajectory segment growth. $n_{t-1}^i$ is the detected target in the last frame $(t-1)$. Under the uniform motion constraint in the short time interval, we define the cost of linking $c_j^t$ to $T^i$. The link involves $n_{t-2}^i, n_{t-1}^i$ and $c_j^t$. Using $n_{t-2}^i$ and $c_j^t$, we get the ideal middle point $n_{t-1}^{i'}$ under the uniform motion constraint. $d_{ij}$ is the Euclidean distance between $n_{t-1}^i$ and $n_{t-1}^{i'}$. The cost $C(i,j)$ of the link is defined as

$$C(i,j) = \frac{d_{ij}}{\|n_{t-2}^i - n_{t-1}^i\|_2} = \frac{\|n_{t-2}^i + c_j^t - 2n_{t-1}^i\|_2}{2 \times \|n_{t-2}^i - n_{t-1}^i\|_2}. \quad (3)$$

The smaller $C(i,j)$ is, the more the link meets the short-strict constraint. The cost matrix $C$ contains all possible links' cost values. We define the binary linking matrix $A_1$ in

$$A_1(i,j) = \begin{cases} 1, & C(i,j) \leq \sigma_1 \\ 0, & C(i,j) > \sigma_1 \end{cases}. \quad (4)$$

where $\sigma_1$ is the cost threshold. We also restrict the absolute velocity value of the target. The restriction for the target candidates $c_j^{t-1}$ and $c_j^t$ in successive frames is defined in (5). $\sigma_2$ is an absolute velocity threshold.

$$CoV(c_j^{t-1}, c_i^t) = \begin{cases} 1, & \|c_j^{t-1} - c_i^t\|_2 \leq \sigma_2 \\ 0, & \|c_j^{t-1} - c_i^t\|_2 > \sigma_2 \end{cases}. \quad (5)$$

For an existing trajectory segment, if more than one target candidate meets (3) and (4), we link the trajectory segment to each target candidate and record every new link as a new trajectory segment. In order to find a new target, we also link the target candidates $\{c_j^t\}_N$ in the current frame $t$ and the candidates $\{c_i^{t-1}\}_{N'}$ in the last frame $(t\text{-}1)$ under (5) to form new trajectory segments. Trajectory segment growth with the short-strict constraint is performed as presented in Algorithm 1. It should be noted that we treat a reentry target as a new target.

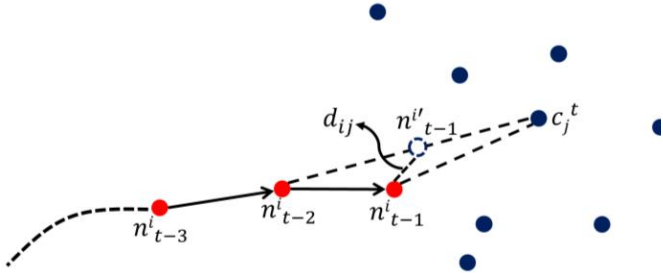

Fig. 3. Trajectory segment growth with short-strict constraint.

| **Algorithm 1:** Trajectory segment growth with the short-strict constraint | **Algorithm 2:** Trajectory segment merging with the long-loose constraint |
|---|---|
| **Input:** $\{T^i\}_M$ —existing trajectory segment set, $\{c_j^t\}_N$ —target candidate set in frame $t$ $\{c_i^{t-1}\}_{N'}$ —target candidate set in frame $(t\text{-}1)$ **Output:** $\{T^i\}_{M'}$ —updated trajectory segment set **for** $i = 1$ *to* $M$ **do**   $flag = 0$   **for** $j = 1$ **to** $N$ **do**     Calculate $A_1(i,j)$ using Equation (3~4)     **if** $A_1(i,j)$ & $CoV(n_{t-1}^i, c_j^t)$         $[T^i, c_j^t] \to \{T^i\}_{M'}$ ; $flag = 1$     **end if**   **end for**   **if** !*flag*     $T^i$ stop growth, $T^i \to \{T^i\}_{M'}$   **end if** **end for** **for** $i = 1$ **to** $N'$ **do**   **for** $j = 1$ **to** $N$ **do**     **if** $CoV(c_i^{t-1}, c_j^t)$         $[c_i^{t-1}, c_j^t] \to \{T^s\}_{M'}$     **end if**   **end for** **end for** | **Input:** $\{T^i\}_M$ —existing trajectory segment set, $A_2$       —binary link matrix **Output:** $\{T^i\}_{M'}$ —updated trajectory segment set **for** $i = 1$ *to* $M$ **do**   $flag = 0$   **for** $j = i+1$ **to** $M$ **do**     **if** $A_2(i,j)$       merge $T^i$ and $T^j$ to $T^{i'}$, $T^{i'} \to \{T^i\}_{M'}$,       $flag = 1$       **for** k = 1 **to** $M$ **do**         **if** $A_2(k,j)$           $A_2(i,k) = A_2(k,i) = 0$         **end if**         **if** $A_2(k,i)$           $A_2(j,k) = A_2(k,j) = 0$         **end if**       **end for**     **end if**   **end for**   **if** !*flag*     $T^i \to \{T^i\}_{M'}$   **end if** **end for** |

*2) Trajectory Segment Merging With Long-Loose Constraint:* As given above, we get a trajectory segment set $\{T^i\}_M$, which meets the piecewise uniform motion short-strict constraint. Given noise interference or a cluttered background, the true target may not be detected correctly from each frame in Section IV-A. Therefore, the true target trajectory is divided into several segments. Trajectory segment merging intends to link the trajectory segments corresponding to the same target. The merging is performed based on the similarity between trajectory segments. Fig. 4 presents three trajectory segment pairs with different relative positions. Compared to the segments in Fig. 4(a) and (c), the segments in Fig. 4(b) are more likely to correspond to the same target.

The features of the trajectory segments corresponding to the same target are as follows.

(1) The trajectory segments do not overlap in time.

(2) The extension of the trajectory segments is close to each other.

(3) The velocity values of different trajectory segments are close to each other.

We take two track segments $T^1$ and $T^2$ as samples to detail the definition of the similarity measure, as shown in Fig. 5.

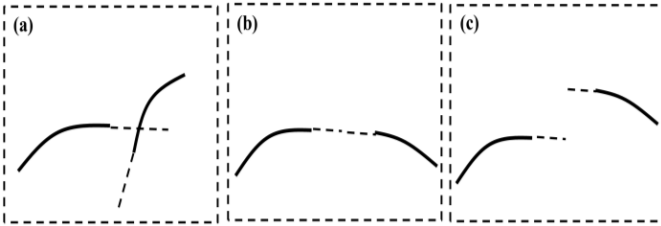

Fig. 4. Trajectory segment pairs with different relative positions. (The dot line represents the extension of the trajectory segment.)

$T^1$ and $T^2$ do not overlap in time. $T^1$ ends at frame $(t-4)$ and $T^2$ starts from frame $(t-1)$. According to the uniform motion constraint, we extend $T^1$ and $T^2$ to frame $(t-3)$ and $(t-2)$ respectively. The extended target positions are $\{p_{t-3}^1, p_{t-2}^1\}$ and $\{p_{t-3}^2, p_{t-2}^2\}$ respectively. If $T^1$ and $T^2$ belong to the same target trajectory, we link $n_{t-4}^1$ and $n_{t-1}^2$.

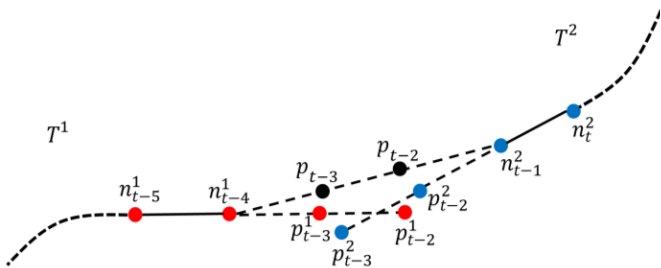

Fig. 5. Similarity measure for trajectory segment merging.

Under the uniform motion constraint, the interpolated target positions are $\{p_{t-3}, p_{t-2}\}$ as shown in Fig. 5. The distances between the extended target positions and the interpolated target positions are used to define the similarity measure $s(T^1, T^2)$ between $T^1$ and $T^2$ as in

$$s(T^1, T^2) = \frac{(t-1)-(t-4)-1}{\sum_{k=t-3}^{t-2}\left(\left\|p_k^1 - p_k\right\|_2 + \left\|p_k^2 - p_k\right\|_2\right)}. \quad (6)$$

The equation above is a detailed description of the similarity definition of two trajectory segments $T^1$ and $T^2$ with a time interval 2. For other cases, the similarity is calculated in a manner similar to the above definition. The similarity matrix $S = \left[s(T^i, T^j)\right]_{M \times M}$ is a symmetric matrix ($S(i,j) = S(j,i)$) with zero diagonal elements ($S(i,i) = 0$). The binary link matrix $A_2$ is given in

$$A_2(i,j) = \begin{cases} 1, & S(i,j) \geq \sigma_3 \\ 0, & S(i,j) < \sigma_3 \end{cases}. \quad (7)$$

where $\sigma_3$ is a threshold to be set. The parameter corresponds to the degree of relaxation of the long-loose constraint. Considering the continuous target movement, we prioritize merging long trajectory segments. We sort the trajectory segments according to their lengths in descending order, i.e., the first row $A_2$ corresponds to the longest trajectory segment in $\{T^i\}_M$. Details of the trajectory segment merging are described in Algorithm 2. We delete trajectory segment showing no growth or merging in the last $\sigma_4$ frames from the list.

After trajectory segment growth and merging, we detect the small infrared aerial target based on the length of the trajectory. The length threshold $\sigma_5$ is defined as

$$\sigma_5 = \lfloor \mu \times L \rfloor. \quad (8)$$

where $\mu$ is a constant and $L$ is the length of the time window. If the length of a trajectory is larger than $\sigma_5$, the corresponding target candidates are detected as true targets. The target can be detected continuously through the trajectory segment growth. We assume that there is only one single target in a position at one time. Thus for the crossed trajectory segments, we keep the longest trajectory and eliminate others.

## V. SMALL AERIAL TARGET DETECTION FOR AIRBORNE INFRARED DETECTION SYSTEM DATASET

A high-quality dataset is essential to promote the development of research, e.g., ImageNet and MS-COCO for object recognition and segmentation. However, to the best of our knowledge, there is no public dataset dedicated to this research so far. For small infrared target detection, the lack of available images is a serious concern, worthy of research attention. In general, researchers expend considerable effort in collecting images for algorithm evaluation and comparison. Experimental images have the disadvantages of small scale and disunity. Table II summarizes the scales of images used in recently published related works [6, 11-15, 19, 36, 38, 40, 43, 44].

It should be noted that Dai *et al*. [43] used their public single frame infrared small target (SIRST) dataset for experimental evaluation. The dataset contains 427 images, including 480 instances of various scenarios. As reported in [43], the SIRST dataset is the largest open dataset for small infrared target detection from a single image at present. However, as given in Table II, several hundred images are not enough to cover the numerous complex situations in real applications. The scale of such datasets are inadequate for learning-based algorithms to train upon, especially for neural networks related algorithms. The lack of test data makes the experimental results unconvincing. Furthermore, the authors of various works collected test images individually, which resulted in disunity among the test images. It is also difficult for researchers to compare their algorithms with other representative algorithms without the released implementation. Furthermore, available public benchmark has not come out at present.

TABLE II
SMALL INFRARED TARGET IMAGES USED FOR EVALUATION AND COMPARISON IN [6, 11-15, 19, 36, 38, 40, 43, 44].

| Algorithm | Images used for evaluation and comparison |
| --- | --- |
| CGF[6] | 12 scenes, 930 images. |
| LAMPS[11] | 165 images. |
| OLD[12] | 5 sequences. |
| Zhou[13] | 10 sequences, 980 images. |
| MIPT[14] | 6 sequences, 1274 images. |
| MGVD[15] | 5 datasets, 495 images. |
| MMP[19] | 6 datasets, 315 images. |
| NMoG-MFD[36] | 5 sequences, 582 images. |
| STTM[38] | 4 synthetic infrared videos and 4 real infrared videos. |
| MSLSTIPT[40] | 6 sequences, 706 images. |
| ALCNet[43] | 427 images: 50% training, 20% validation, and 30% testing. |
| IRSTD-GAN[44] | Training images: 1000 paired synthesized images. Testing images: 8 image sequences. |

*A. Dataset Description*

To meet this need, a large scale dataset, called SIATD, is built for airborne infrared detection systems. The images are captured by airborne infrared imaging equipment. The details of the dataset are presented in Table III.

SIATD contains 350 image sequences, 150185 images, and detailed annotation files. The image sequences are categorized into five scene types. The annotation file provides the target positions in the images. There are at most three targets per frame. The size of the target varies from $3\times3$ pixels to $7\times7$ pixels. The movements of the targets are diverse. To the best of our knowledge, this dataset comprises the largest data scale and richest scene types in this research field. It can be used in single frame as well as multiple frames small infrared target detection studies. The scale of the proposed dataset covers numerous application scenes and provides a solid basis for algorithm research. Specially, for neural networks related algorithms, SIATD provides adequate images for model training.

The dataset contains training and testing subsets. Each contains 175 image sequences. We released the SIATD dataset through a publicly accessible website (http://dx.doi.org/10.11922/sciencedb.j00001.00231). The training subset has been fully released, including the image sequences and annotation files, while only the image sequences in the testing subset have been released. We adopt public used metrics and provide a convenient testing interface on the website. Researchers can upload their detection results on the testing subset in a standard format to the website to evaluate the results. The website will provide a real-time ranking leaderboard accordingly. The dataset provides a public platform for researchers to perform algorithm evaluation and comparison conveniently.

TABLE III
DETAILS OF SIATD DATASET.

| Scene Type | Up looking | | | Head up looking | | | Down looking | | | | | | | | |
| --- | --- | --- | --- | --- | --- | --- | --- | --- | --- | --- | --- | --- | --- | --- | --- |
| | | | | | | | Vegetation | | | Water | | | Building | | |
| Target Number | 1 | 2 | 3 | 1 | 2 | 3 | 1 | 2 | 3 | 1 | 2 | 3 | 1 | 2 | 3 |
| Image sequences | 30 | 20 | 20 | 30 | 20 | 20 | 30 | 20 | 20 | 30 | 20 | 20 | 30 | 20 | 20 |

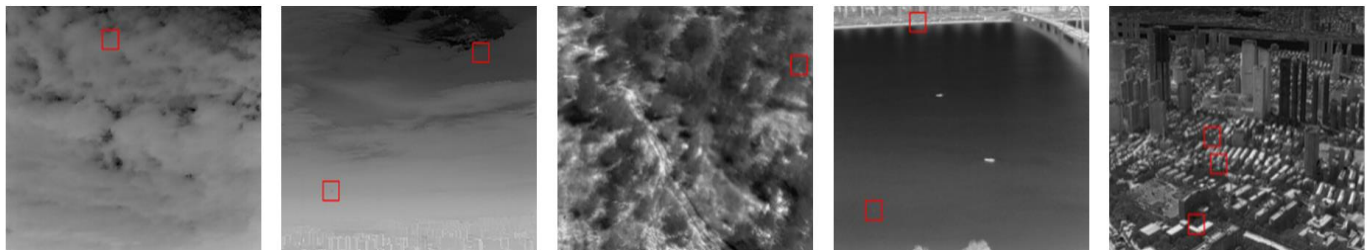

Fig. 6. Sample images picked form the dataset SIATD. (From left to right: Up-looking, Head up looking, Down looking (Vegetation), Down looking (Water), Down looking (Building). The targets are labeled by rectangle.)

As mentioned above, the clutter and rapidly changing of the background bring great difficulties to small moving target detection for airborne infrared detection system. The two main factors are the typical features of SIATD that different from other existing datasets. The typical features make SIATD a more challenge dataset for small aerial infrared target detection.

## B. Evaluation Metrics

In remote imaging applications, dim and small infrared aerial targets are often presented as spot targets in infrared images. As shown Fig. 2 and Fig. 6, they only occupy a few pixels in the images. Given the infrared imaging characteristics, the target edges are blurred. The scarcity of target intrinsic characteristics means that the shape of the target has little significance for practical application. We can simply use the center position of the spot to represent the small infrared aerial target in images. Therefore, we do not use the signal to clutter ratio, background suppression factor, intersection over union, or receiver operating characteristics as metrics, as used in previous works. We adopt a direct and publicly used metric, $F_\beta - measure$, in SIATD to evaluate the algorithm performance quantitatively. The metric combines the precision and recall of the evaluated algorithm. It is defined as

$$F_\beta = \frac{(1+\beta^2) \cdot \text{Precision} \cdot \text{Recall}}{\beta^2 \cdot \text{Precision} + \text{Recall}}$$
$$\text{Precision} = \frac{TP}{TP+FP} \quad . \quad (9)$$
$$\text{Recall} = \frac{TP}{TP+FN}$$

Here, TP, FP, and FN denote the correctly detected target number, wrongly detected target number and the missed target number, respectively. We set $\beta^2 = 1$ in this article.

## VI. EXPERIMENTS AND ANALYSIS

### A. Experimental Settings

To validate the performance of the proposed algorithm qualitatively and quantitatively, we conduct experiments on the public datasets SIRST and SIATD. We also perform comparisons between our method and representative existing methods, including single frame based methods (RIPT[9], MLCM[17], ALCNet[43], AGADM[47]) and multiple frames based method (TIPI[34], ASTTV-NTLA [35], STTV-WNIPT [39]). For the compared methods, we use the implementations released by the authors and their suggested default parameter settings. For the proposed method, the parameter settings used are listed in Table IV in this article, except the experiments conducted in ablation study. Briefly, we denote our target candidate detection algorithm as "Med-LGBM" and the complete method as "proposed method". Target is correctly detected if the location detected is within the three-pixel neighborhood of the ground truth.

All experiments run on a PC with NVIDIA 1080 GPU, i7-8700K CPU and 24GB of RAM. The proposed method is implemented in C++ and CUDA. Using the parameter setting listed in Table IV, it takes about 61 ms per image in average, thereof 44 ms for target candidate detection, and 17 ms for trajectory constraining. Further parallel design can improve the efficiency of the proposed method.

### B. Parameter Sensitivity Analysis

The parameters included in the proposed method and the default settings are summarized in Table IV. We start by investigating the sensitivity of the relatively important

TABLE IV
PARAMETER SETTINGS OF THE PROPOSED METHOD.

| Parameter | Setting |
|---|---|
| Number of target candidates per image | 3500 |
| Sampling parameters in LightGBM [2] | 0.2, 0.3 |
| Local region size $R_1 \times R_2$ | $3\times3, 7\times7, 11\times11, 15\times15$ |
| Link cost threshold $\sigma_1$ | 0.2 |
| Velocity threshold $\sigma_2$ | 10 |
| Trajectory segment similarity threshold $\sigma_3$ | 0.1 |
| Trajectory segment stopped threshold $\sigma_4$ | 5 |
| Length percentage $\mu$ | 0.3 |
| Time window length $L$ | 20 |

parameters: $k_1$, $k_2$, $\mu$ and $L$.

As mentioned above, $k_1$ and $k_2$ are used to select target candidate from per frame. They are determined by varying from both sides to middle according to the required number of selected target candidate in practice. Here we vary the number of target candidates selected from per image and the results on the testing subset of SIATD are reported in Table V.

As shown in Table V, with the increase of the target candidate number per frame, the performance of the proposed method continues to improve. However, the range of improvement gradually decreases. In addition, the increase of the target candidate number will increase the computational burden for the following process. A reasonable target candidate number should be chosen to balance the efficiency and effectiveness of the proposed method in practice.

TABLE V
PERFORMANCE WITH VARYING TARGET CANDIDATE NUMBER ON SIATD.

| Target candidate number | Recall | Precision | $F_\beta - measure$ |
|---|---|---|---|
| 300 | 0.36 | 0.98 | 0.54 |
| 1100 | 0.44 | 0.99 | 0.60 |
| 1900 | 0.46 | 0.99 | 0.62 |
| 2700 | 0.48 | 0.99 | 0.65 |
| 3500 | 0.48 | 0.99 | 0.65 |
| 4300 | 0.50 | 0.99 | 0.67 |
| 5100 | 0.50 | 0.99 | 0.67 |

$\mu$ and $L$ determine the trajectory length threshold ($\lfloor \mu \times L \rfloor$). We set $L=20$ and vary $\mu$ to evaluate the performance of the proposed method with different trajectory length threshold. The results are presented in Table VI.

The results in Table VI indicate that the recall decrease and precision increase with the increase of $\mu$. A larger length threshold means more stringent constraints. Only target candidates with high confidence can be retained when using larger $\mu$. However, the comprehensive metric $F_\beta - measure$ deceases while $\mu$ increases. The main reason is that many true targets are removed by the larger trajectory threshold. Due to missed detection of targets in adjacent frames, some correctly detected targets cannot form long continuous trajectories.

TABLE VI
PERFORMANCE WITH VARYING $\mu$ ON SIATD.

| $\mu$ | Recall | Precision | $F_\beta - measure$ |
|---|---|---|---|
| 0.15 | 0.59 | 0.90 | 0.72 |
| 0.25 | 0.57 | 0.94 | 0.71 |
| 0.35 | 0.55 | 0.97 | 0.70 |
| 0.45 | 0.48 | 0.99 | 0.65 |
| 0.55 | 0.45 | 1.00 | 0.62 |
| 0.65 | 0.42 | 1.00 | 0.59 |
| 0.75 | 0.39 | 1.00 | 0.56 |
| 0.85 | 0.36 | 1.00 | 0.53 |

*C. Target Candidate Detection From Single Image*

Med-LGBM detects the target candidates and suppresses the background interferences through interesting pixel detection and a trained LightGBM model. Actually, all existing single-frame-based algorithms can be embedded in our framework to detect the target candidate from a single frame. With trajectory constraints, the detection of the true target is based on an underlying premise that it is first determined as a target candidate. To ensure the effectiveness of small infrared aerial target detection, we require the target candidate detection to be as effective as possible. High recall means that the true targets are likely to be detected and high precision means fewer false alarms. The lower the number of target candidates is, the more efficient the subsequent trajectory segment growth and merging will be. We compare the proposed target candidate detection method with representative single-frame-based methods on SIRST. For the methods that output target region, e.g., ALCNet, we take the geometric center of the detected target region as the final detection result. Fig. 7 shows sample results of the target candidate detection. It is noted that only the correct detections are labeled for the compared methods. Owing to too many false alarms detected by some compared methods (e.g., the third - fifth row in Fig. 7), the false detections are left unlabeled. However, as Med-LGBM does not have too much false detection, we label them in presented results.

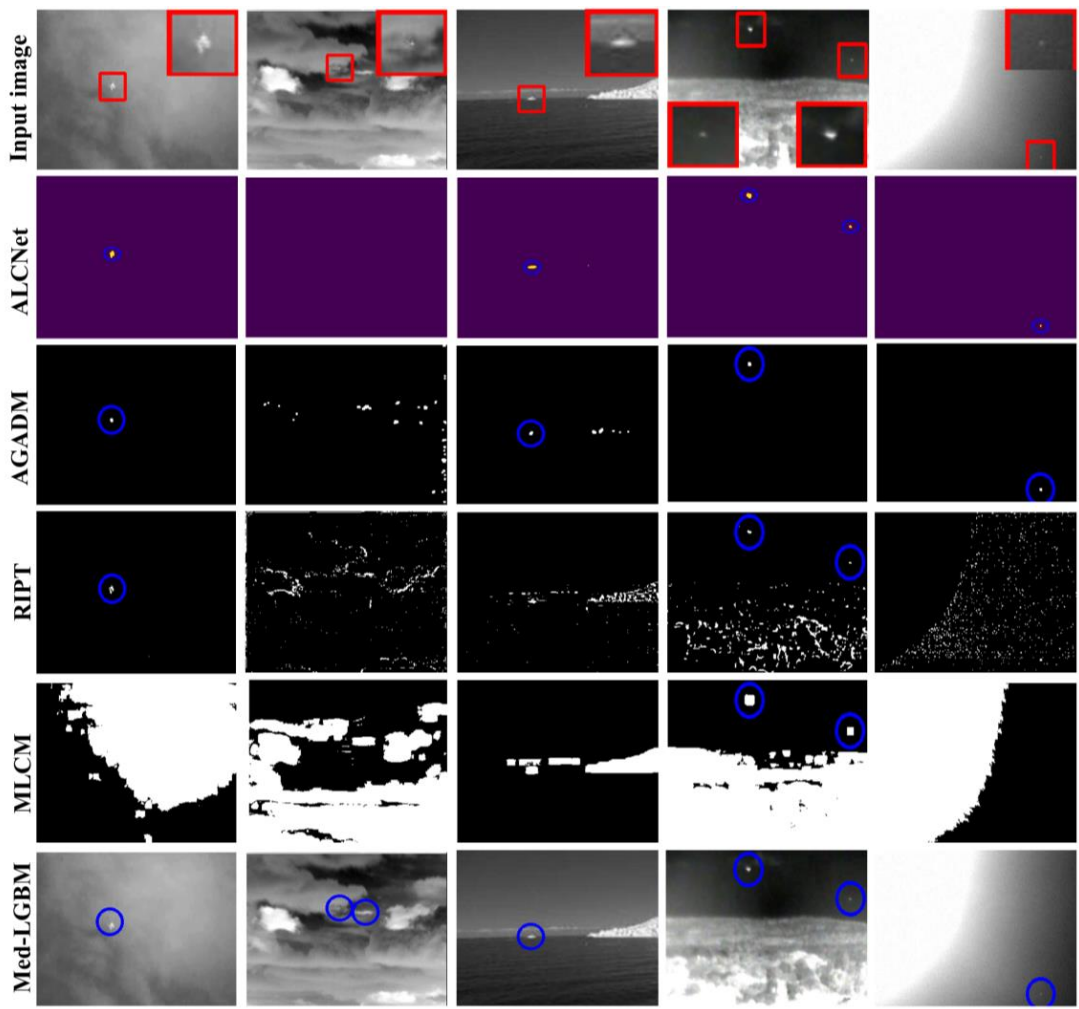

Fig. 7. Sample results of the target candidate detection from SIRST. (The detected targets are labeled by "O" and the true targets by "□".)

The results demonstrate that most of the true targets are correctly detected by ALCNet and Med-LGBM. ALCNet and AGADM have difficulties in detecting small dim targets as the second image in Fig. 7 shows. The cluttered background presents challenges for MLCM and RIPT.

The statistical results on SIRST are listed in Table VII. We compute the geometric center of the label mask for each labeled target as the ground truth. The recall and precision are calculated as given in Section V-B. The results in Table VII demonstrated that Med-LGBM achieves slightly lower performance than ALCNet and performs better than other methods. The high recall indicates that most true targets are detected as target candidates. The high precision of Med-LGBM makes efficient trajectory growth and merging possible. RIPT achieves the highest recall. However, the low precision causes certain difficulties. It should be noted that Med-LGBM aims at a small-sized spot target. Some targets in SIRST are of large size and do not satisfy the hypothesis of a small spot target. This causes the performance of Med-LGBM to suffer.

ALCNet is a CNN-based method and needs sufficient training data. As reported in [43], the SIRST dataset only contains 427 images. We adopt the ALCNet model and Med-LGBM trained on the SIRST dataset to detect small infrared aerial targets in images in the testing subset from SIATD dataset. The image in SIATD is of size $640\times512$ pixels. ALCNet resizes the input image to a standard size of $256\times256$ pixels. The direct resize operation of the image from SIATD may make the small target even smaller. For fairness, we crop the test images manually and only preserve the local region of size $256\times256$ pixels containing the target. Sample detection results of ALCNet and Med-LGBM are presented in Fig. 8. The statistical results are given in Table VIII. It should be noted that the LightGBM model used in Med-LGBM is also trained on the SIRST dataset.

In general, the targets in SIATD are smaller than those in SIRST. The results in Fig. 8 and Table VIII indicate that Med-LGBM achieves better performance than ALCNet on images picked from SIATD, especially for small dim targets. ALCNet also outputs false detections corresponding to cluttered backgrounds (see the third - fifth column in Fig. 8). ALCNet may need more training data to improve its ability compared to other data-driven methods. Med-LGBM has better generalization ability than ALCNet, as the experimental results show.

TABLE VII
STATISTICS OF THE TARGET CANDIDATE DETECTION ON SIRST.

| Algorithm | Recall | Precision | $F_\beta$ – measure |
|---|---|---|---|
| ALCNet | 0.98 | 0.93 | 0.95 |
| AGADM | 0.88 | 0.39 | 0.55 |
| RIPT | 0.98 | 0.005 | 0.02 |
| MLCM | 0.86 | 0.1 | 0.18 |
| Med-LGBM | 0.90 | 0.77 | 0.82 |

TABLE VIII
STATISTICS OF THE TARGET CANDIDATE DETECTION ON IMAGES PICKED FROM SIATD.

| Algorithm | Recall | Precision | $F_\beta$ – measure |
|---|---|---|---|
| ALCNet | 0.08 | 0.15 | 0.10 |
| Med-LGBM | 0.32 | 0.19 | 0.24 |

### D. Target Detection From Image Sequence

In this section, we conduct experiments of supon image sequences in the SIATD dataset. The proposed method first detects target candidates from each individual frame using interesting pixel detection and a trained LightGBM model as described in Section IV-A. For successive frames, a simple commonly-used target candidate detection method is inter-frame differencing and thresholding. We compare the simple method with the proposed target candidate detection method on the testing subset of the SIATD dataset. For the inter-frame-differencing-based method, we perform inter-frame registration as described in Section IV-B. Threshold is determined adaptively via Otsu [48]. The LightGBM model is trained on the training subset of the SIATD dataset. The results are presented in Table IX.

The results in Table IX indicate that the proposed target candidate detection method achieves higher recall and precision than the inter-frame-differencing-based method. Approximately 0.87 target candidates on average are detected by the proposed method, which makes the trajectory growth and merging more efficient. It should be noted that there are at most three true targets in each frame and the targets may move out of view in SIATD. Thus, the average number of target candidates per frame of the proposed target candidate detection method is less than one. The low average number indicates the high precision of the proposed target candidate detection method to a certain extent.

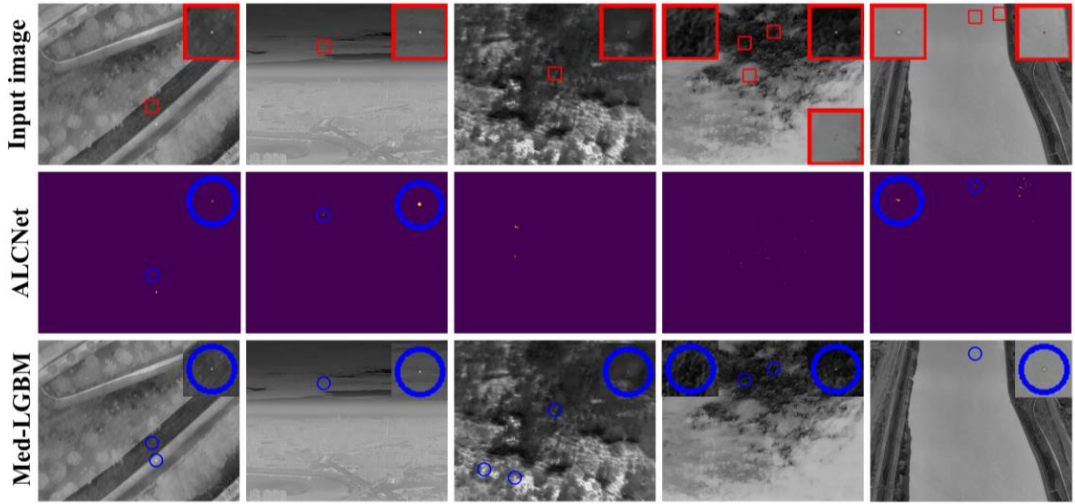

Fig. 8. Sample results of the target candidate detection on images picked from SIATD. (The detected target candidates are labeled by "○" and the true targets by "□".)

TABLE IX
TARGET CANDIDATE DETECTION ON THE TESTING SUBSET OF SIATD DATASET.

| Algorithm | Recall | Precision | $F_\beta$ – measure | Average number of target candidate per frame |
|---|---|---|---|---|
| Inter-frame differencing | 0.73 | 0.30 | 0.43 | 3.96 |
| Med-LGBM | 0.74 | 0.84 | 0.79 | 0.87 |

The proposed method detects the true targets from target candidates using trajectory constraints. The short-strict and long-loose constraints described in Section IV-B enable the proposed method to track long target trajectories. We present two detected trajectories within two sample image sequences from SIATD in Fig. 9. The detected targets in the sequence are remapped to the same coordinate as in Fig. 2. It can be seen from Fig. 9 that the targets' trajectories are tortuous yet smooth and continuous. The proposed method detects them correctly. Using the trajectory constraints, the clutter within images is eliminated effectively by our method.

It should be noted that the LightGBM model used in the proposed method needs training using annotated data, while the compared methods in this experiment do not. We train the LightGBM model using the training subset in SIATD. To be fair, we only report the results on the testing subset in this section. ALCNet needs pixel-wise labeled target region as training data while SIATD only provides the labeled target center position. Furthermore, we have proved in Section IV-B that ALCNet has trouble in detecting extremely small dim infrared aerial targets in SIATD. Therefore, this study does not consider ALCNet as a comparison method in this experiment. Fig. 10 shows sample detection results from the SIATD dataset. As shown in Fig. 7, we only label the correct detections for the compared methods.

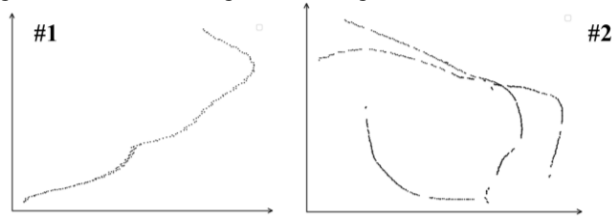

Fig. 9. The detected trajectories for two sample image sequences form SIATD dataset.

The results in Fig. 10 show that existing algorithms have trouble in detecting small infrared aerial targets, especially for targets under cluttered backgrounds. Clutter causes considerable difficulties for AGADM, RIPT, and MLCM. False targets are detected in cluttered background areas as shown in Fig. 10. TIPI has trouble in modeling the quick change in the background. Strong edges cause false detections, as shown in the fifth row in Fig. 10. In the latest published works, STTV-WNIPT and ASTTV-NTLA adopt more advanced technics in reconstruction. However, they still have trouble in detecting infrared small target under clutter and changing backgrounds, as shown in the fifth and seventh row in Fig. 10. The proposed method detects the targets accurately and performs better than other algorithms. As shown in the second column in Fig. 10, a darker target is located in the building region. The proposed method detects it correctly while others can not.

For quantitative evaluation, the quantitative evaluations of each algorithm on the testing subset are reported in Table VIII, including results of each scene type and the entire testing subset.

As mentioned above, cluttered backgrounds bring considerable challenges for existing single frame based methods. AGADM, RIPT, and MLCM achieve lower precisions than the proposed method. The clutter degrees of the backgrounds within the Down looking scene are higher than that within the Up looking scene generally. The performances of the compared methods decrease with the increase of the clutter degree, as presented in Table X. Our target candidate detection method Med-LGBM achieves better performance than them. LightGBM is a learning-based method. Its performance heavily depends on the training data. We reported the results for each scene type separately and the entire testing subset in Table X. The results on the entire testing subset are similar to those on various scene types. This indicates that Med-LGBM has the ability to deal with a variety of complex scenes.

TIPI cannot handle a quick-changing cluttered background satisfactorily. Consequently, it achieves poor performance on the SIATD dataset. As illustrated in Fig. 10, rapidly changing and clutter backgrounds also bring great challenges to STTV-WNIPT and ASTTV-NTLA. They also achieve poorer performances than the proposed method, especially for down looking (vegetation) and down looking (building) scene types. By introducing the short-strict and long-loose trajectory constraints, the proposed method eliminates false detections and improves the precision significantly. The proposed method achieves better performance than existing algorithms in detecting small infrared aerial targets.

The results in Table X indicate that the precision is improved considerably from Med-LGBM to the proposed method by introducing trajectory constrains. False detections are effectively removed. In the image sequence, unlike the true target, the clutter within the background cannot form smooth and continuous trajectories. However, as Table X presents, the recall decreases from Med- LGBM to the proposed method. This means that some correctly detected targets are removed by mistake during trajectory growth and merging. We analyzed the experimental results and found that most of the mistaken removed target candidates are isolated detected targets. There is no detected target close to them in the preceding and succeeding frames. Therefore, they cannot form valid trajectory segments and are removed in the final detection.

Although the proposed method achieves superior performance than existing methods in the above experiments, it still has many limitations. The clutter background brings great troubles to Med-LGBM. Missing or false detection will avoid the true target to be correctly detected, e.g. a true target is missing in the leftmost column in Fg.10. In result, the recalls of the proposed method reported in Table X are low. The performance can be further improved by exploring more advanced target candidate detection methods and linking categories. We believe that neural networks related achievements in spatial and temporal data processing have immense potential for improving SIATD.

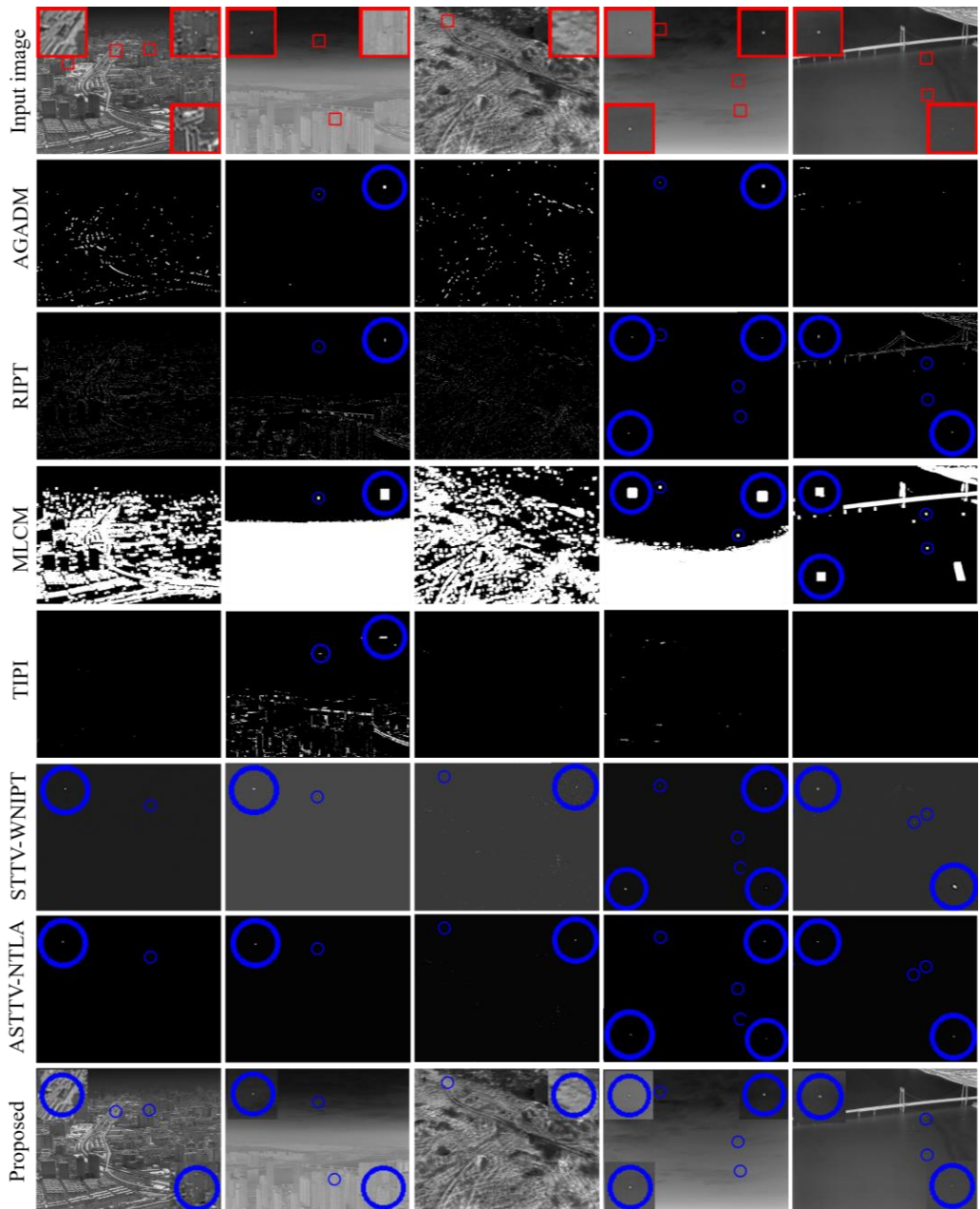

Fig. 10. Sample detection results of AGADM, RIPT, MLCM, TIPI, STTV-WNIPT, ASTTV-NTLA and the proposed method from SIATD. (The detected target candidates are labeled by "O" and the true targets by "□".)

TABLE X
QUANTITATIVE EVALUATION OF AGADM, RIPT, MLCM, TIPI, STTV-WNIPT, ASTTV-NTLA AND THE PROPOSED METHOD ON THE TESTING SUBSET OF SIATD. (**U** FOR UP LOOKING, **H** FOR HEAD UP LOOKING, **D-V** FOR DOWN LOOKING (VEGETATION), **D-W** FOR DOWN LOOKING (WATER), **D-B** FOR DOWN LOOKING (BUILDING), **P** FOR PRECISION, **R** FOR RECALL AND **F** FOR $F_\beta - measure$ .)

|  | AGADM | | | RIPT | | | MLCM | | | TIPI | | | Med-LGBM | | | STTV-WNIPT | | | ASTTV-NTLA | | | Proposed | | |
|---|---|---|---|---|---|---|---|---|---|---|---|---|---|---|---|---|---|---|---|---|---|---|---|---|
|  | P | R | F | P | R | F | P | R | F | P | R | F | P | R | F | P | R | F | P | R | F | P | R | F |
| U | 0.14 | 0.33 | 0.20 | 0.33 | 0.79 | 0.47 | 0.26 | 0.61 | 0.36 | 0.04 | 0.37 | 0.07 | 0.54 | 0.87 | 0.67 | 0.91 | 0.43 | 0.59 | 0.91 | 0.44 | 0.59 | 0.95 | 0.66 | 0.78 |
| H | 0.14 | 0.33 | 0.20 | 0.33 | 0.79 | 0.47 | 0.26 | 0.61 | 0.36 | 0.01 | 0.02 | 0.01. | 0.42 | 0.85 | 0.56 | 0.13 | 0.40 | 0.20 | 0.44 | 0.37 | 0.40 | 0.76 | 0.65 | 0.74 |
| D-V | 0.03 | 0.18 | 0.06 | 0.26 | 0.63 | 0.38 | 0.14 | 0.33 | 0.20 | 0.01 | 0.01 | 0.01 | 0.87 | 0.72 | 0.79 | 0.67 | 0.62 | 0.64 | 0.99 | 0.56 | 0.71 | 0.99 | 0.60 | 0.75 |
| D-W | 0.01 | 0.07 | 0.02 | 0.32 | 0.78 | 0.46 | 0.13 | 0.33 | 0.19 | 0.01 | 0.02 | 0.01 | 0.50 | 0.88 | 0.64 | 0.99 | 0.60 | 0.75 | 0.99 | 0.60 | 0.75 | 0.93 | 0.69 | 0.79 |
| D-B | 0.01 | 0.05 | 0.02 | 0.30 | 0.74 | 0.44 | 0.13 | 0.33 | 0.20 | 0.00 | 0.00 | 0.00 | 0.68 | 0.59 | 0.64 | 0.18 | 0.53 | 0.27 | 0.34 | 0.62 | 0.44 | 0.98 | 0.57 | 0.72 |
| Whole | 0.01 | 0.03 | 0.01 | 0.30 | 0.75 | 0.43 | 0.05 | 0.13 | 0.08 | 0.01 | 0.09 | 0.02 | 0.68 | 0.76 | 0.72 | 0.21 | 0.50 | 0.30 | 0.60 | 0.52 | 0.56 | 0.92 | 0.63 | 0.75 |

## VII. CONCLUSIONS

This article focuses on the challenge of small aerial target detection for airborne infrared detection systems. We first analyze the characteristics of the target trajectory and point out that the basic characteristics of the target trajectory, continuous and smooth, are of immense significance for robust SIATD. A simple and effective SIATD method using LightGBM and trajectory constraints is then proposed. Target candidate detection from each individual frame using the spatial cuesis treated as a binary classification problem. Interesting pixel detection and a trained LightGBM model are applied to detect target candidates. Multiple spatial features are extracted and inputted to the LightGBM model. We adopt the piecewise uniform motion model to approximate the target movement. The true targets are detected from amongst the target candidates using trajectory constraints, including the short-strict and long-loose constraints. The constraints are used in trajectory segment growth and merging. Experiments on publicly available datasets indicate that the proposed method detects small infrared aerial target robustly and achieves better performance than existing methods. In order to alleviate the lacuna between research needs and the availability of testing data, we also build a high-quality SIATD dataset and release it to the public. To the best of our knowledge, our dataset possesses the largest data scale and the richest scene types in this field at present.

For future work, neural networks related achievements in spatial and temporal data processing can be harnessed in SIATD. The powerful feature extraction and representation ability of neural networks has immense potential to improve the performance of SIATD.

## ACKNOWLEDGMENT

Special thanks to Max Halford, Ting Liu *et al*. for their selfless sharing.

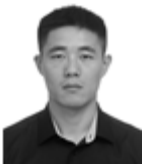
XIAOLIANG SUN received the B.S., M.S., and Ph.D. degrees from the National University of Defense Technology, Changsha, China, in 2010, 2013, and 2017, respectively. He is currently a Lecturer with the College of Aerospace Science and Engineering, National University of Defense Technology. His main research interests include image measurement, machine vision, and navigation systems.

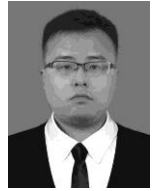
LIANGCHAO GUO received the B.S. degree from Xiangtan University (XTU), Xiangtan, China, in 2018. He is currently pursuing the M.S. degree with the College of Aerospace Science and Engineering in National University of Defense Technology. His research interests include infrared target detection and image enhancement.

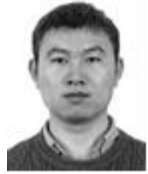
WEMLONG ZHANG received the B.S., M.S. and Ph.D. degrees from the National University of Defense Technology, Changsha, Chine, in 2014, 2016 and 2020, respectively. He is currently a lecturer with the College of aerospace science and engineering, National University of Defense Technology. His research interests include image processing and compute vision.

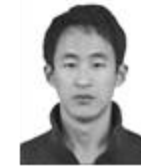
ZI WANG received the B.S. degree from Tianjin University, Tianjin, China, in 2016, and the M.S. degree from the National University of Defense Technology, Changsha, China, in 2018, where he is currently pursuing the Ph.D. degree with the College of Aerospace Science and Engineering. His research interests include deep learning, object pose detection, and computer vision.

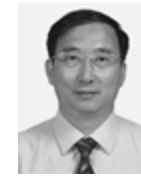
QIFENG YU received the B.S. degree from Northwestern Polytechnic University, Xi'an, China, in 1981, the M.S. degree from the National University of Defense Technology, Changsha, China, in 1984, and the Ph.D. degree from Bremen University, Bremen, Germany, in 1996. He is currently a Professor with the National University of Defense Technology. He has authored three books and published over 100 articles. His main research fields are image measurement, vision navigation, and close-range photogrammetry.